# Blockchain and Artificial Intelligence

Tshilidzi Marwala and Bo Xing

University of Johannesburg

Auckland Park

Republic of South Africa

**Abstract:** It is undeniable that artificial intelligence (AI) and blockchain concepts are spreading at a phenomenal rate. Both technologies have distinct degree of technological complexity and multi-dimensional business implications. However, a common misunderstanding about blockchain concept, in particular, is that "a blockchain is decentralized and therefore no one controls it". But the underlying development of a blockchain system is still attributed to a cluster of core developers. Take smart contract as an example, it is essentially a collection of codes (or functions) and data (or states) that are programmed and deployed on a blockchain (say, Ethereum) by different human programmers. It is thus, unfortunately, less likely to be free of loopholes and flaws. In this article, through a brief overview about how artificial intelligence could be used to deliver bug-free smart contract so as to achieve the goal of blockchain 2.0, we to emphasize that the blockchain implementation can be assisted or enhanced via various AI techniques. The alliance of AI and blockchain is expected to create numerous possibilities.

## 1. Introduction

Nowadays, one can hardly ignore the role of artificial intelligence (AI) and blockchain in the wave of the fourth industrial revolution (4IR): the former is integrated into 4IR's DNA and the latter could revolutionize economic system's infrastructure. We believe the joint force of these two technologies could determine the depth and breadth of the 4IR. In order to discuss the synergy of AI and blockchain, it is important to understand what AI and blockchain are in the first place.

## 1.1 What is Artificial Intelligence?

Intelligence is the ability to make sense of information beyond the obvious. There are two types of intelligence in nature and these are individual intelligence and group intelligence [1]. As the name implies, AI is made out of two words, artificial and intelligence and thus it is intelligence that is artificially made. Various types of artificial intelligence techniques have been proposed and these include neural networks, support vector machines and fuzzy logic. These techniques have been successfully used for missing data estimation, finite element models, modelling interstate conflict, economics and in robotics.

## 1.2 What is Blockchain?

In the simplest term, blockchain denotes an unalterable digital ledger system. One notable feature of blockchain technology is its distributed implementation manner. It initially originated from Bitcoin, which has now demonstrated its potential in numerous domains. Further discussions regarding blockchain can be found in many publications, e.g., [2].

## 2. A Snapshot of Smart Contract Security Issue

Software or software parts are widely found in numerous devices and gadgets. It is thus not exaggerated to say that the well and correct functioning of our society relies largely on software systems. Under this circumstance, the dependability and reliability of any (either embedded or commercial) software system play a crucial role in determining success. Recently, with the popularity of blockchain concept, smart contract alike hold the promise of further decentralizing our market environments in many ways.

Nevertheless, at present, the defined standards regarding smart contract security is still lacking which means the hidden vulnerabilities found in new or existing smart contracts could potentially lead to undesired results, e.g., monetary losses. In 2017 alone, the hack of

Ethereum's Parity wallet caused a total loss of $180 million; while in 2016, a bug found in

Ethereum's DAO (i.e., decentralized autonomous organization) enabled a hacker to siphon $50 million from its smart contract [3].

### 2.1 Smart Contract

In principle, a smart contract can perform various operations automatically [4], e.g., calculations, information storage, funds transfer, etc. In JavaScript terms [5], one can view a contract as a class which consists of different elements such as state variables, functions, function modifiers, events, structures, and enums [6]. According to [6], smart contracts can also support inheritance and polymorphism. In a simple case, a programmer can store the hash of a file and the owner's name as pairs in the code to achieve the functionality of proof of ownership. Similarly, one can also store the hash of a file and the block's timestamp as pairs to realize the proof of existence function. Finally, by storing the hash itself, the file integrity can be proven, i.e., if the file is altered, the hash will change correspondingly and smart contract is then incapable of finding any such file.

### 2.2 Blockchain 2.0

According to [2], blockchain 2.0 represents the next key step of blockchain development. Since the space of blockchain 2.0 is still under construction, some emerging concepts belonging to this category may include smart contract, smart asset, decentralized autonomous organizations, and so on.

### 2.3 AI Assisted Smart Contract Testing

The aim of this section is to explore how smart contract testing can be fulfilled by employing AI techniques. Accordingly, the following points, illustrated in Fig. 1, serve as the roadmap for our discussion.

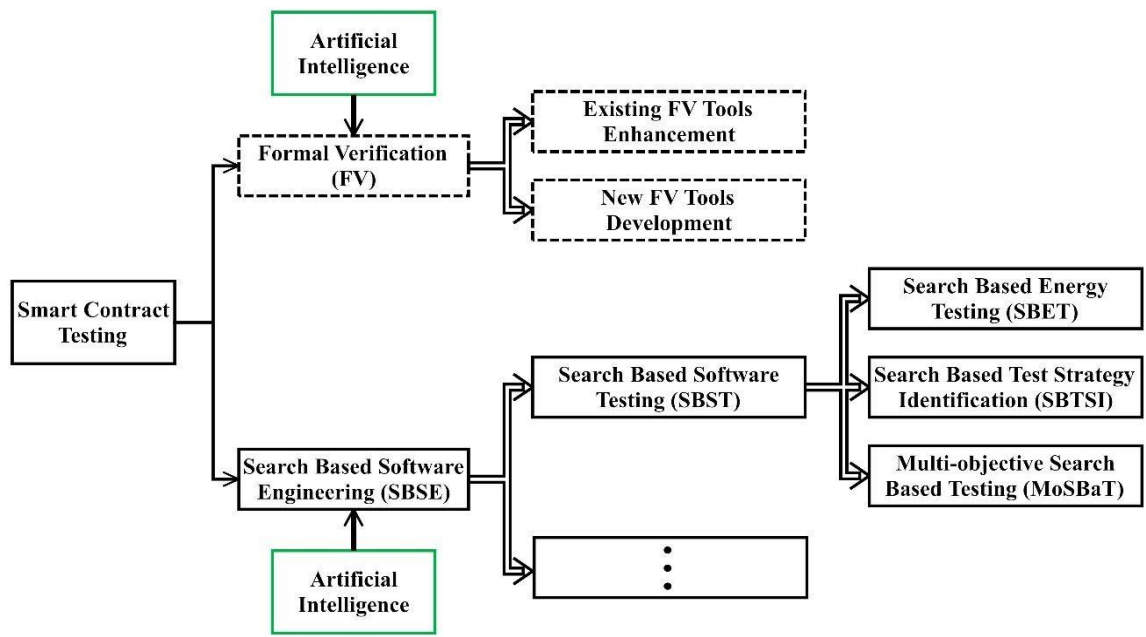

**Figure 1:**    **Two Main Testing Categories.**

### 2.3.1  Formal Verification

In its simplest term, formal verification (FV) stands for the utilization of mathematical reasoning to ensure computing systems' correct execution [7]. Generally speaking the applications of AI in formal verification can be classified as follows:

- Broad Domains:
  - ➢ Enhance the existing FV tools; and ➢    Facilitate new FV tools development.
- Applications areas:
  - ➢ Automate troubleshooting;
  - ➢ Debugging and root cause analysis and identification;
  - ➢ Assist theorem proving; and
  - ➢ Learning a concept from a concept class.

For instance, since writing specifications manually tends to be complicated and error prone, AI can learn specifications from runtime traces. In addition, AI can be used to select heuristic strategy automatically, that is, choosing heuristic according to need-to-prove conjecture's features and the corresponding axioms.

### 2.3.2 Search Based Software Engineering

In principle, search based software engineering (SBSE) utilizes computational search methodologies to address various software engineering problems which are often characterized by their large complex search spaces. As a sub-area of software engineering, thought the origins of SBSE can be stretched all the way back to the 1970s, it was until 2001, SBSE was formally established as a field of study. Since roughly half of the total funding spent on software projects goes to software testing [8], it is thus not surprising that over half of SBSE publications are also dedicated to search based software testing (SBST). For more details in this regard, please refer to [9-14].

### 2.3.3 Summary

Security is one of the major concerns that is currently holding smart contract concept back from mainstream adoption. Unlike its counterpart centralized systems, once a smart contract is put into practice on a decentralized blockchain, rollbacks and compensations are often hard to be performed when coding errors occur. Though human security audit could be a solution for smart contract, for creators lift their level of security, the cost associated with this practice tend to be a discouraging factor. This section briefly explores the role of AI in facilitating smart contract testing. Though such applications are still rare in the literature, the authors believe the possibility and feasibility space for employing AI techniques, e.g., [15, 16], to verify smart contract is huge.

## 3. AI and Blockchain: A United Approach

The focus of this article is to show how AI can help us implement blockchain technology.

Accordingly, the follow aspects, illustrated in Fig. 2, serve as the starting point of synthesizing AI and blockchain. These aspects were identified by Corea [17].

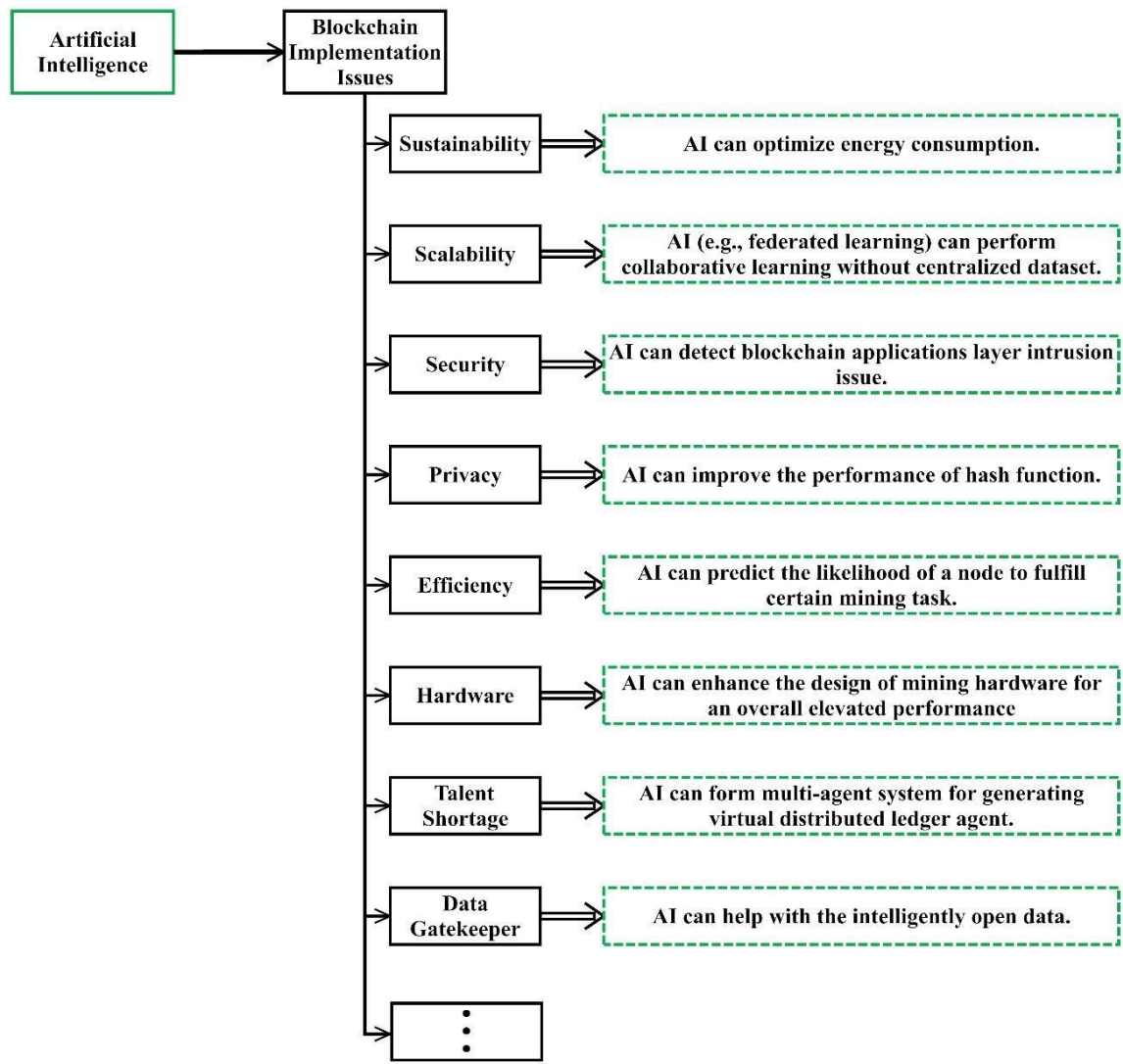

**Figure 2:**   **The Synergy of AI and Blockchain.**

### 3.1 Sustainability

The methodologies of AI have long been applied to optimize large-scale system (e.g., power system planning and operation). On the other hand, intelligent optimization algorithms are also the basic tools for analysing microeconomic environment. Essentially, blockchain (or distributed ledger system) and microeconomics both are large-scale distributed systems, and there are inherent connections found between them. In principle, a blockchain system

(including various nodes such as full node, mining node, and lightweight node) and a microeconomic system (including a social system, which consists of producers, consumers, and markets) have many similarities: different interconnected subsystems, decentralized computations, and so on. The broad concern of microeconomics is to allocate scarce resources among various uses with the goal of maximizing users' utility and producers' profit. Then a unified view of AI-backed blockchain system energy consumption optimization can be established from the large-scale complex systems perspective [18].

### 3.2 Scalability

Scalability in the context of blockchain generally refers to its scale capability with the increase of the number of users. In practice, we can view scalability issues via various angles such as latency (the time required for confirming a transaction), bootstrap time (the time consumed for validating a transaction), and cost generated per confirmed transactions. Overall, the efficiency of a blockchain system is limited by one or more of these scalability issues. Since each block contains a certain amount of transaction data, conventional centralized data mining techniques are struggling to cope with this situation. Nevertheless, novel AI algorithms (say, federated learning) can learn from distributed data sources, which in turn offer us a global optimal solution for the target blockchain system.

### 3.3 Security

The security concerns of a blockchain system cover the applications layer vulnerability (say, smart contract), data encryption mechanism, etc. In terms of the vulnerability applications layer, the intrusion detection system (IDS) and intrusion protection system (IPS) are critical components for monitoring various threats. In order to increase the efficacy of an IDS, swarm intelligence [15] (a sub-branch of AI which seeks inspiration from swarms of different biological system) approaches have been widely applied in this direction. Regarding blockchain data encryption mechanism, computational intelligence (another key branch area of AI) also plays an important role in both classical and modern cryptographic systems. Their applications in this regard range from cryptography (e.g., cellular automata and DNA computing), cryptanalysis (e.g., evolutionary computation), and hash function (artificial neural network). In general, the advantages of using computational intelligence include creating more robust ciphers and improving blockchain system's resilience via computational intelligence improved system attack-defence process.

### 3.4 Privacy

With more and more personal data embedded in blockchain system, the data encryption becomes a prominent issue for guaranteeing users' privacy. This aspect is more or less related to previous security issue in which we have showed the important role of AI. Take Bitcoin blockchain system, it is currently utilizing elliptic curves based private and public key generation. However, for now, no one has managed to develop a public-key algorithm that is free from weaknesses. To better address this problem, different intelligent search algorithms can be used collectively for searching the bits of a secret key (an in-depth exploration and exploitation of the search space).

### 3.5 Efficiency

In blockchain network, simply getting the total throughput maximized is not always sufficient for maintaining a desired transaction validation performance. Take a sensor network, when we use it to track certain objects' mobility in a large observation space, the total throughput maximization only strategy might cause fairness issue among different mining nodes (i.e., majority nodes could be excluded due to data transportation cost). Network utility maximization (NUM) model can assist us in seeking distributed solution for controlling congestion, routing, and scheduling in computer networks (including Internet and emerging blockchain network). Since NUM is essentially a twice differentiable utility function that is characterized by its concavity and non-decreasing features, and most importantly in many practical scenarios, the amount of available resources is unknown a priori. AI can, therefore, perform active and dynamic learning so as to accelerate resources estimation and improve overall system's performance.

### 3.6 Hardware

Specialized computer components (mainly supplied by Shenzhen, China) plays a crucial role in keeping a blockchain system running. The current computer architecture is mainly built on von Neumann architecture which classified a computer into different components such as central processing unit (CPU), internal memory, external storage, input/output (I/O) devices, and buses (wires used to connect these components together). Other computer architectures also include Harvard architecture, RISC (reduced instruction set computer), and parallel processing architecture. In this regard, neural-inspired neuromorphic hardware [19, 20] shed light on a new direction. One example of such hardware design could consist of a couple of hundred neurons along with numerous synaptic phase change memory cells that are built on leaky integrate-and-fire and spike-timing-dependent plasticity spiking neuron models.

### 3.7 Talent Shortage

Given the current short supply of blockchain personnel, one possible way is to introduce multi-agent approach. By creating various task-oriented virtual agent, the process of writing/reading transaction data from blocks can thus be fully automated. On the other hand, AI-assisted online learning can also help us train and nurture much needed blockchain talent.

### 3.8 Data Gatekeeper

With the wide spread of data economy, an intelligent open data is a top priority. As data resources built on blockchain technology become more and more available, both companies and individuals need help with the data at hand regarding their accessibility, utilization, and sense-making. The power of AI makes it perfectly suitable for this type of tasks.

## 4. Conclusion

The DNA of the 4IR is AI, while blockchain represents one of the most disruptive technologies that may transform the whole economic system. Though blockchain holds various promises, this technology is still in its infancy. In this article, we have outlined several blockchain related implementation concerns. Based on our preliminary research, we also highlighted possible solution avenue(s) through the lens of AI. It is the authors' hope that this article inspire other researchers from diverse disciplines to bring the application of AI in blockchain domain to its full potential.